%% file: paper_v2.tex
\documentclass[11pt]{article}
\usepackage[final]{acl}
\usepackage{times}
\usepackage{latexsym}
\usepackage[T1]{fontenc}
\usepackage[utf8]{inputenc}
\usepackage{microtype}
\usepackage{graphicx}
\usepackage{booktabs}
\usepackage{url}

\title{Decomposition-Induced Context-Memory Conflict: \\
An Instruction-Induced Failure and a Probing Case Study}

\author{Yu-Feng Yen \\ Independent Researcher \\ \texttt{ccuhollis@alum.ccu.edu.tw}}

\begin{document}
\maketitle

\begin{abstract}
Decompose-then-verify pipelines (FActScore-style fact-checkers, hallucination detectors) first split a passage into atomic claims and treat this step as neutral preprocessing. We ask whether a decomposer can substitute its own parametric belief for what the passage says, producing a claim that contradicts its source, and call this \emph{Decomposition-Induced Context-Memory Conflict} (DI-CC). On biographies with a deliberately altered year, Qwen2.5-7B-Instruct does so in 4.3\% of claims when the instruction licenses self-correction, but in only 0.2--0.5\% under a neutral instruction, a FActScore-style few-shot prompt, a ``use only the passage'' instruction, a formatting instruction, or a knowledge-filling instruction (an upper bound: in the neutral and FActScore-style runs only 1 of 13 flagged claims is a real override); context-aware decoding removes only half. DI-CC here is therefore instruction-induced. We then test whether a linear probe trained only on classical context-memory conflict data (NQ-Swap) detects DI-CC zero-shot. Its AUC is high (up to 0.94 against legit claims, 0.98 under a position- and instruction-matched control), but random directions reach AUCs between about 0.1 and 0.9 on these contrasts, probes fit directly on the claims reach 0.96--1.00 everywhere (even on a control that separates only the text variant), and after multiplicity correction the NQ-Swap probe beats random directions only at the last layer, and only for one of two random-direction families. A direct NLI check between source and claim reaches AUC 0.998. High probe AUC on such small contrasts is not evidence of a shared mechanism without nulls, and two pitfalls, pair leakage in probe training and duplicated BOS tokens, can change such conclusions.
\end{abstract}

\section{Introduction}

Long-form factuality evaluation increasingly follows a \emph{decompose-then-verify} recipe: split a generated passage into atomic claims, then check each against a reference \citep{min2023factscore,wanner2024dndscore}. Decomposition is treated as bookkeeping. The decomposer is itself a language model, however, and language models tend to substitute parametric knowledge for context when the two disagree, a phenomenon studied as \emph{context-memory conflict} \citep{longpre2021entity}. Prior work examines this at final-answer generation. We ask whether it also occurs \emph{inside decomposition}, so that a claim later ``verified'' against the source contradicts that very source.

We call this \textbf{DI-CC}: a decomposer, given source passage $R$, emits an atomic claim $c_i$ that (a) is not entailed by $R$, (b) contains content recoverable from the decomposer's own closed-book knowledge, and (c) directly contradicts $R$.

Behavioral evidence alone cannot show that DI-CC is the \emph{same phenomenon} as classical context-memory conflict rather than an unrelated hallucination that looks similar. We therefore ask a mechanistic question: if the two share a mechanism, a probe trained \emph{exclusively} on classical conflict data should detect DI-CC positions zero-shot. We test this with controls chosen to separate conflict from confounds (the instruction, the position, and the amount of added content) and with null probes that show what AUC an arbitrary direction obtains.

\paragraph{Scope.} The DI-CC we measure is largely \emph{elicited}: our baseline instruction explicitly licenses the model to check facts against its own knowledge. Under a neutral instruction and a FActScore-style prompt it falls to 0.2--0.3\% of claims (Section~\ref{sec:cad}), although 2--3\% of year-bearing claims still restate the original year. Our evidence therefore concerns how the failure behaves under controls, not how often it occurs in deployed pipelines. We also did not measure its downstream effect on verification scores, where a silently corrected claim could mask a hallucination in the source.

\paragraph{Contributions.}
(1) We define DI-CC and show that it is induced by a correction-licensing instruction: neutral, FActScore-style and passage-only instructions leave 0.2--0.5\% of claims, and a passage-only instruction outperforms context-aware decoding. (2) We test whether an NQ-Swap probe transfers to DI-CC with entity-clustered statistics, a matched control, null probes and in-domain reference probes, and show that high AUCs are mostly indistinguishable from arbitrary directions and far below in-domain probes: after correction only the last layer beats random directions, and only under one of two ways of drawing them. (3) We report two pitfalls in our own probing pipeline, pair leakage in probe training and BOS duplication.

\section{Problem Definition}

Let $D$ be a decomposer LLM with parametric knowledge $M_D$, and $c_i$ the $i$-th atomic claim of $D(R)$. Claim $c_i$ exhibits \emph{parametric injection} if its content $I=\mathrm{content}(c_i)\setminus\mathrm{content}(R)$ satisfies $R\not\models c_i$ (judged by an NLI model) and $I\in\mathrm{Recoverable}(M_D)$ (closed-book probing of $D$ without $R$ reproduces $I$). \textbf{DI-CC}: $I$ contradicts information in $R$. A non-contradictory sibling, unsupported elaboration (DI-UE), fills a position $R$ never addressed. Unsupported claims whose content is not recoverable from $M_D$ are excluded.

\paragraph{Question.} If DI-CC is mechanistically related to classical context-memory conflict, a probe $\mathbf{w}$ trained only on classical conflict data (NQ-Swap) and applied zero-shot to decomposition activations should separate DI-CC positions from legit ones. The layer-sweep probing method follows \citet{pham2026where,zhao2024spare}.

\section{Related Work}

\paragraph{Knowledge conflict.} NQ-Swap \citep{longpre2021entity} replaces a gold answer entity with a same-type distractor so context and closed-book answer disagree; context-aware decoding (CAD) contrasts output distributions with and without context \citep{shi2023cad}; ConflictBank \citep{su2024conflictbank} scales this up; \citet{xu2024knowledgeconflicts} survey the area. Prompting is also an established way to raise context faithfulness \citep{zhou2023contextfaithful}, and mechanistic work separates recall of a fact from copying an in-context counterfactual \citep{ortu2024competition,jin2024cutting}. All of this studies conflict at answer generation, not decomposition; our passage-only-instruction result is a decomposition-specific instance of the prompting finding, and our matched control contrasts recall with copying, the two mechanisms those papers study.

\paragraph{Decomposition and detection.} FActScore \citep{min2023factscore} established atomic decompose-then-verify. DnDScore \citep{wanner2024dndscore} and CREDENCE \citep{tran2026credence} note that decomposition can introduce error, but neither formalizes a conflict-specific failure or tests for a shared mechanism. SelfCheckGPT \citep{manakul2023selfcheckgpt} flags hallucination via cross-sample disagreement.

\paragraph{Layer selection and controls for probes.} Control tasks and selectivity \citep{hewitt2019control} separate what a probe reads out from what it can fit; our null and in-domain probes play the analogous role for AUC on small contrasts. Accuracy-selected layers can transfer poorly out of distribution \citep{uselis2025intermediate}, and the between/within-class variance ratio is an alternative criterion \citep{burger2024truth}. Linear truth structure is preserved within, but not necessarily across, model lineages \citep{choi2026truth,zolfaghari2026linear,moschella2023relative}. We add a case where the layer, and the null used to judge it, decide the whole conclusion.

\section{Setup}
\label{sec:setup}

\paragraph{Models.} The decomposer and probe host is Qwen2.5-7B-Instruct (4-bit NF4), decoded greedily unless stated. Entailment and recoverability use one NLI model (DeBERTa-v3-large, MNLI/FEVER/ANLI/LingNLI/WANLI).

\paragraph{Texts and instructions.} Source texts are self-generated closed-book biographies of 191 well-known people, in a \emph{vague} (pronoun-heavy) and a \emph{perturbed} variant (a year shifted NQ-Swap-style); 176 entities have a perturbed biography (the other 15 biographies contain no year to alter). On the perturbed biographies we compare seven decomposition instructions. \textbf{Licensing} (our baseline) asks the model to check each claim against what it knows and output the corrected version of any incorrect fact. \textbf{Passage-only} asks it to check against the biography only and not to correct. \textbf{Sham} is a length-matched instruction about formatting. \textbf{Knowledge-filling} asks it to replace vague phrases with specific details it knows (the instruction used for vague biographies). \textbf{Neutral} asks only for atomic claims with pronouns resolved. \textbf{FActScore-style} splits the biography into sentences and prompts the model with eight fixed few-shot demonstrations from the FActScore repository \citep{min2023factscore}, read as a plain continuation; it is not the original implementation, which used a different model and added one retrieved demonstration. \textbf{CAD} uses the licensing instruction with context-aware decoding ($\alpha{=}1.0$) \citep{shi2023cad}.

\paragraph{Labels.} Each claim is labeled legit, DI-CC, DI-UE or excluded by NLI between the source, the claim and the model's closed-book dump. For checks that must not depend on this NLI model, we use a \emph{string label} on perturbed biographies: a claim containing the original year is an \emph{override}, one containing the altered year is a \emph{copy}. The \emph{matched control} is the contrast between override and copy claims (108 versus 62) under the licensing instruction: same text, same instruction, same position type, and the model either recalls the true year or copies the stated one.

\begin{figure}[t]
\centering
\includegraphics[width=\columnwidth]{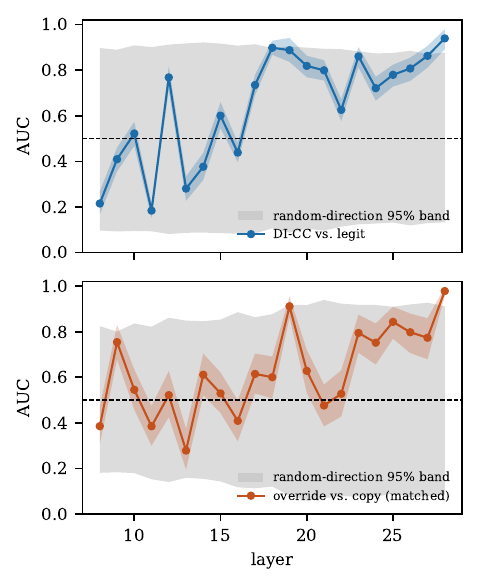}
\caption{AUC by layer of the NQ-Swap probe on decomposition activations. Top: DI-CC vs.\ legit; bottom: override vs.\ copy (matched control). Colored band: 95\% entity-clustered bootstrap CI. Gray band: 2.5th--97.5th percentile of 1,000 random directions. Dashed: chance.}
\label{fig:layers}
\end{figure}

\paragraph{Probe and nulls.} A logistic regression is trained on 852 NQ-Swap activations (conflict and no-conflict pairs) at the final prompt token, separately for each hidden layer, and applied zero-shot at the claim-end token of each decomposition. We split the training data \emph{by example}, so the two versions of one example are never separated (Section~\ref{sec:pitfalls}). To judge an AUC we score the same claims with null probes at every layer: 1,000 isotropic Gaussian directions, and 30 probes trained on NQ-Swap with shuffled labels, and compare against their 2.5th--97.5th percentiles (these draws give Figure~\ref{fig:layers} and Table~\ref{tab:layers}). For corrected p-values we draw fresh nulls of 2,000 directions each from three families: isotropic Gaussian, Gaussian with the covariance of the claim activations, and Gaussian with the covariance of the NQ-Swap training activations (plus 100 shuffled-label probes). As reference points we fit a logistic-regression probe and a difference-of-means direction directly on the claims (5-fold, folds grouped by entity), and apply the NQ-Swap difference-of-means direction zero-shot.

\paragraph{Statistics.} AUC intervals use an entity-clustered bootstrap (2000 resamples). The 67 positives of the pooled DI-CC-versus-legit set (62 from perturbed and 5 from vague biographies) lie in 63 entities, so clustering barely changes them (DI-CC vs.\ legit, layer 20 with the earlier row-level split: cluster [0.866, 0.939], i.i.d.\ [0.863, 0.940]). A max-statistic label permutation over layers 8--28 (2000 permutations) tests whether any layer separates the labels. Empirical p-values against the nulls are $(1+\#\{\mathrm{null}\ge\mathrm{obs}\})/(K+1)$, two-sided on $|\mathrm{AUC}-0.5|$, with Holm correction over the 21 layers of each contrast.

\paragraph{Provenance.} The intermediate files of an earlier run of this pipeline were lost, so every number below comes from a complete regeneration with the released code: one greedy run, plus the seed experiments in Section~\ref{sec:probe}. Applying the probes saved from the earlier run to the regenerated activations gives AUC 0.875 and 0.839 at layer 20 (DI-CC vs.\ legit and vs.\ DI-UE) and 0.837 and 0.867 at layer 22, within 0.03 of the earlier run's values (released reports), so we treat the regenerated data as a close replacement. Counts are not identical (67 versus 64 positives).

\section{Results}

\subsection{DI-CC is instruction-induced and mitigable}
\label{sec:cad}

\begin{table}[t]
\centering\footnotesize
\begin{tabular}{@{}lrrrr@{}}
\toprule
Instruction & Claims & DI-CC & \% [95\% CI] & Entities \\
\midrule
Licensing (baseline) & 1,449 & 62 & 4.3 [3.4, 5.2] & 60 \\
Neutral & 1,748 & 3 & 0.2 [0.0, 0.4] & 3 \\
FActScore-style$^\dagger$ & 3,082 & 10 & 0.3 [0.1, 0.7] & 6 \\
Knowledge-filling & 1,744 & 9 & 0.5 [0.2, 0.9] & 9 \\
Passage-only & 1,799 & 3 & 0.2 [0.0, 0.4] & 3 \\
Sham (length-matched) & 1,730 & 5 & 0.3 [0.1, 0.6] & 5 \\
CAD ($\alpha{=}1.0$) & 1,356 & 31 & 2.3 [1.5, 3.1] & 29 \\
\bottomrule
\end{tabular}
\caption{DI-CC on 176 perturbed biographies (NLI labels, so rows other than licensing and CAD are upper bounds; see text). Entities: number of entities with at least one DI-CC claim (of 176). CI: entity-clustered bootstrap. Exact McNemar on entity incidence versus the baseline: $p{<}10^{-13}$ for every row except CAD ($p{=}3\times10^{-6}$). $^\dagger$Approximation of the published prompt (Setup); it decomposes sentence by sentence, so it has more claims and is not directly comparable to the other rows.}
\label{tab:cad}
\end{table}

DI-CC appears almost only when the instruction licenses correcting the text (Table~\ref{tab:cad}). The neutral, FActScore-style, passage-only, formatting and knowledge-filling instructions leave 3, 10, 3, 5 and 9 cases (0.2--0.5\%), and the altered year is copied in nearly all year-bearing claims (for knowledge-filling, 172 of 183). A small number still restate the original year: 4 of 186 year-bearing claims under the neutral instruction and 5 of 189 under the FActScore-style prompt (2--3\%), against 108 of 170 under licensing. The passage-only and sham instructions also yield more claims than the baseline (1,799 and 1,730 versus 1,449), so the reduction is not an output-length effect. We inspected the flagged claims against the passages (one annotator, not blind). Of the 3 passage-only and 5 sham cases, 5 are overrides of very well-known facts (the year of Hillary's summit, the birth years of Serena Williams and Muhammad Ali) and 3 are NLI mislabels of claims the passage supports. Of the 3 neutral cases, one is a real override (Serena Williams's birth year) and two are claims the passage states. Of the 10 FActScore-style cases, one is a real override (Hillary's summit date); the other nine are decomposition artifacts, not parametric injection: sentence fragments with a lost subject (``Louis, Missouri, was an influential poet'', four times; ``Seventh Army showcased leadership'', twice), a pronoun fragment, a misattribution (Hardy credited with studying at Cambridge) and a copied instruction line. The true override rate under these two prompts is therefore about one claim in 1,700 and 3,100, and the percentages in Table~\ref{tab:cad} are upper bounds. Conversely, of the 108 licensing-condition claims containing the original year, NLI labels 55 as DI-CC and 50 as excluded, so the DI-CC counts are conservative.

CAD roughly halves DI-CC (62 to 31) and shortens output by 6.4\%. Dropping 6.4\% of baseline claims at random leaves 31.9\% of entities affected (95\% interval 29.5--33.5\%), against 34.1\% at baseline and 16.5\% under CAD, so CAD's reduction exceeds the effect of shorter output. Because a neutral or passage-only instruction already removes most of it at no cost, we treat CAD as a demonstration that the mechanism can be manipulated, not as the recommended mitigation.

\subsection{Probe transfer: high AUC, wide nulls}
\label{sec:probe}

\begin{table}[t]
\centering\scriptsize
\setlength{\tabcolsep}{2pt}
\begin{tabular}{@{}lccccc@{}}
\toprule
Layer & DI-CC vs.\ legit & Matched & Cond.\ ID & Null$_{97.5}$ & sd (D / M) \\
\midrule
\input{tab_layers.tex}
\end{tabular}
\caption{AUC [95\% CI] of the NQ-Swap probe. Matched: override versus copy of the altered year (108 vs.\ 62 claims, string labels). Cond.\ ID: perturbed versus vague among legit claims. Null$_{97.5}$: 97.5th percentile of the random-direction AUC on the matched control. sd: standard deviation of the AUC over five temperature-0.7 regenerations of all decompositions, for DI-CC vs.\ legit (D) and matched (M). Full 21-layer table in Appendix~\ref{app:layers}.}
\label{tab:layers}
\end{table}

\paragraph{AUC is high at many layers.} Across layers 8--28 the maximum AUC for DI-CC versus legit claims is 0.94 and for the matched control 0.98 (Table~\ref{tab:layers}); the max-statistic permutation $p$ is at its floor (0.0005) for both. The AUC depends strongly on the layer: layer 28 gives 0.98 on the matched control and layer 22 gives 0.53 (difference 0.45, 95\% CI [0.35, 0.55]). At a fixed layer the AUC is stable over five regenerations of every decomposition at temperature 0.7 (standard deviation 0.004--0.025) and over 20 probe-training splits (0.01--0.10), while layer means span 0.19--0.96 and 0.29--0.97 for the two contrasts. This shows that the data and the probe training are stable; it is not evidence of signal, because any fixed direction has a fixed AUC at each layer (next paragraph).

\paragraph{But the null is wide.} Random directions score between 0.09 and 0.91 on the matched control at layer 28 (95\% band), and the 95\% bands are 0.6--0.9 wide at every layer; the bands are wider still when directions follow the claim-activation covariance (below), and high-variance activation directions plausibly separate these small, structured contrasts; we did not test this. Many real-probe AUCs are accordingly indistinguishable from arbitrary directions, and the layers where the probe is significantly \emph{below} chance (layer 8: 0.22; layer 11: 0.18) fall inside the null band as well. Without correction the probe exceeds the random-direction band at layers 18 and 28 for DI-CC vs.\ legit, and at layer 28 on the matched control; it exceeds the shuffled-label band at 5 and 4 of the 21 layers (DI-CC vs.\ legit: layers 18, 23, 25, 27, 28; matched: layers 19, 23, 25, 28). Layers 23 and 25, which have a high matched AUC and a condition-identity AUC near chance (0.47, 0.50), exceed only the shuffled-label band and fall inside the random-direction band, so we do not read them as signal. Under the criterion we fixed before running the nulls (the probe must exceed them at at least half of the layers), the claim that most layers carry a transferable conflict signal is not supported.

\paragraph{After multiplicity correction.} We compare against 2,000 random directions with Holm correction over the 21 layers, drawn in three ways. With isotropic directions only layer 28 is significant, on both contrasts (uncorrected $p=0.001$, Holm-adjusted 0.03 and 0.02); layer 18 (uncorrected $p=0.046$, adjusted 0.93) is not. With directions drawn from the covariance of the NQ-Swap training activations the result is the same (adjusted 0.03 and 0.01). With directions drawn from the covariance of the claim activations, which makes the null bands even wider (0.66--0.94), no layer is significant (layer 28: uncorrected $p=0.009$ and 0.007, adjusted 0.19 and 0.15). The one significant layer therefore depends on how arbitrary directions are drawn. The shuffled-label null (100 draws) cannot reach significance after correction (its smallest attainable $p$ is about 0.01), so at this resolution it is uninformative.

\paragraph{In-domain reference points.} A probe fit directly on the claims reaches AUC 0.96--0.98 for DI-CC vs.\ legit, 0.98--1.00 on the matched control and 1.00 on the condition-identity control at every layer, and a difference-of-means direction fit the same way reaches 0.86--1.00. All three contrasts are therefore nearly trivially separable by a linear direction, including the one that separates only the text variant. The NQ-Swap probe sits far below this ceiling at almost every layer, and the NQ-Swap difference-of-means direction applied zero-shot ranges from 0.19 to 0.92, as erratically as the logistic probe.

\paragraph{A conflict-free copy-versus-recall contrast.} To see whether the probe merely separates copying from recalling, we score a contrast with no source conflict: within the vague condition (same instruction, same text variant), legit claims that restate the source (308) against DI-UE claims that add recalled detail the source never states (157, 153 entities). The NQ-Swap probe reaches at most 0.70 (layers 17 and 20) and 0.665 at layer 28, where it exceeds the isotropic and claim-covariance bands uncorrected ($p=0.03$, $0.02$) but not after Holm correction (best adjusted $p$ 0.61 for isotropic and 0.39 for claim-covariance directions); an in-domain probe reaches 0.93--0.96 on the same claims. The layer-28 AUCs on the conflict contrasts (0.94 and 0.98) are therefore well above what a recall-versus-copy contrast gives, which argues against the simplest alternative, although the two contrasts differ in content type and claim count, so the comparison is indicative only.

\paragraph{Layer 28 and the condition confound.} Layer 28 is the last hidden state (indices follow HuggingFace \texttt{hidden\_states}, 0 being the embedding; in the library version we used, the last entry is replaced by the model output after the final RMSNorm, which we confirmed in the source, and the mean activation norm falls from 393 at layer 27 to 273 at layer 28, so layer 28 is not on the same scale as the others). It is the only layer significant after correction under two of the three null families. Its condition-identity control (a probe separating perturbed from vague legit claims) is also high there (0.76, inside the null band), and at layer 19 the matched control (0.91) comes with a condition identity of 0.86 that exceeds both nulls before correction, so some of the apparent signal may encode the instruction or text variant. The layer-28 AUC on the matched control is 0.99 for claims that end in the year token (22 vs.\ 39) and 0.98 for claims where the year is not last (86 vs.\ 23), so it is not driven by the final token being a year. We have not tested whether it reflects next-token prediction rather than conflict. The DI-CC versus DI-UE contrast is not usable for specificity: 157 of the 158 DI-UE claims come from the vague condition, so it is almost entirely a condition contrast.

\paragraph{No layer rule is reliable.} The earlier protocol selects the layer by held-out NQ-Swap accuracy or by variance ratio, and with its row-level split both select layer 20, where the matched control is 0.63 and inside the null band. Accuracy does not point to the layers with the strongest transfer (mean held-out accuracy over 20 by-example splits is 0.88 at layer 20 and 0.82 at layer 28).

\subsection{A direct NLI check is not informative here}

On the string-labeled matched set, NLI contradiction probability between the source and the claim reaches AUC 0.998, against 0.63 for the probe at layer 20 (0.98 at layer 28). Because the string labels are almost readable by an NLI model by construction, this shows that the contrast is easy, not that the probe is weak. Against NLI-defined labels the comparison is circular, so we use the string labels. SelfCheckGPT \citep{manakul2023selfcheckgpt} was at chance in the earlier run (0.508 [0.453, 0.564], released reports), as expected: DI-CC content is stable under resampling by construction.

\subsection{Two pitfalls in probe analyses}
\label{sec:pitfalls}

\paragraph{Pair leakage.} NQ-Swap examples are stored as pairs (conflict, no-conflict) of one example. A random row-level split puts one of a pair in training and the other in test, so the test row is a near-copy of a training row with the opposite label. Held-out accuracy at layer 8 is then 0.244 (below chance); splitting by example gives 0.56, and the best-layer accuracy is about 0.88. Held-out accuracy under a leaky split is not a usable layer-selection criterion.

\paragraph{Duplicated BOS.} Tokenizing the output of a chat template (which already inserts BOS) a second time duplicates the BOS token for models such as Mistral. On Mistral-7B-Instruct-v0.3, with the pipeline's original protocol (layer 18, 76 DI-CC positives) the probe is at chance (AUC 0.534, CI [0.48, 0.59]); the released probe gives the same value as a retrained one. After the fix, layer selection and the AUC change (82 positives); we computed no null bands for Mistral, so we do not interpret the fixed result and report it only to show that tokenization handling changes the conclusion.

\section{What We Do Not Rely On}
\label{sec:bounds}

An earlier run of this pipeline reported a below-chance Mistral result (AUC 0.397, released reports) that did not reproduce (Section~\ref{sec:pitfalls}), and a minimum model size for transfer from a 3B, 7B and 14B ablation selected at a single layer, which our layer analysis shows to be unreliable; we withdraw both. Other measurements of that run (natural injection rate, behavior on FActScore hallucinations, agreement with human labels) cannot be checked against retained data and are omitted.

\section{Conclusion}

A decomposer can overwrite its source with its own beliefs when its instruction licenses correction; neutral, FActScore-style and passage-only instructions leave 0.2--0.3\% of claims, so DI-CC is better understood as instruction following on a conflicting source than as a spontaneous failure. A probe trained only on classical conflict data reaches high AUC on these positions at many layers of Qwen2.5-7B, but random directions reach comparable AUCs, probes fit on the claims reach near-perfect AUC on every contrast, and only the last layer beats random directions after correction, and only for two of three ways of drawing them; there it is also partly confounded with the instruction. We recommend that probing studies on small contrasts report the full layer curve, a matched control, null probes and an in-domain reference.

\section*{Limitations}
\label{sec:limits}

\begin{itemize}\setlength\itemsep{0pt}
\item \textbf{Elicited setting.} All DI-CC is elicited by a licensing instruction on self-generated biographies of well-known people; it does not measure natural frequency, and a natural rate on unlabeled text was not re-measured. The FActScore-style prompt approximates the original (fixed demonstrations, a chat model read as a completion model), and we did not measure the downstream effect on verification scores.
\item \textbf{One family.} Positive results use Qwen2.5-7B (4-bit), one NLI model and year perturbations of self-generated biographies of well-known people; Mistral results depend on layer and tokenization and have no null analysis.
\item \textbf{Regenerated data.} The earlier run's intermediate data were lost; results are one regeneration plus seed experiments (temperature 0.7, perturbed condition only), and counts differ slightly from the earlier run.
\item \textbf{Nulls.} The null probes cover arbitrary (Gaussian) directions and one structured alternative (recall versus copy among vague-condition claims, 157 versus 308, with a different content type); the one significant layer depends on the covariance used to draw the arbitrary directions; the shuffled-label null has 30 draws per layer (100 for the corrected p-values). The matched control contrasts recall with copying and does not separate conflict from that contrast in general; it covers only year-bearing claims. The probe is trained at the last prompt token and applied at the claim-end token, a position shift we did not analyze.
\item \textbf{Labels.} The knowledge dump is not a truth oracle, single-digit numeric differences can fool the NLI check, and we inspected residual cases with a single non-blind annotator and did not sample claims labeled legit.
\item \textbf{Mitigation.} The CAD comparison uses one $\alpha$ and does not combine CAD with the passage-only instruction; a parse-failure problem under CAD on vague biographies, seen in the earlier run, was not regenerated.
\end{itemize}

\section*{Ethical Considerations}
All data are model-generated biographies of public figures plus public benchmarks (NQ-Swap, FActScore); no personal or private data were collected. Licenses are listed in Appendix~\ref{app:license}. Detecting model unfaithfulness is a defensive goal; we make no deployment claim.


\bibliography{refs}

\appendix

\section{Full Layer Table}
\label{app:layers}
AUC of the NQ-Swap probe (by-example split) for every layer on the greedy run: DI-CC vs.\ legit, matched override vs.\ copy, condition identity.

\begin{table}[h]
\centering\footnotesize
\begin{tabular}{@{}lccc@{}}
\toprule
Layer & DI-CC vs.\ legit & Matched & Cond.\ ID \\
\midrule
\input{tab_layers_full.tex}
\end{tabular}
\end{table}

\section{Licenses and Compute}
\label{app:license}
Qwen2.5-7B-Instruct and Mistral-7B-Instruct-v0.3 are Apache 2.0. NQ-Swap is under a custom Apple license permitting use and redistribution; FActScore is MIT. Our code and data are MIT-licensed, at \url{https://github.com/hollis-png/di-cc-decomposition-conflict}. The regeneration ran on one NVIDIA RTX 5090 (32GB) shared with other services; a full decomposition, labeling and extraction pass over the perturbed condition takes 7--12 minutes, and the null-probe analysis about 1.5 hours on CPU.

\end{document}

%% file: tab_layers.tex
18 & 0.90 [0.87,0.93] & 0.60 [0.51,0.69] & 0.52 & 0.88 & 0.007 / 0.024 \\
19 & 0.89 [0.84,0.94] & 0.91 [0.86,0.95] & 0.86 & 0.92 & 0.016 / 0.010 \\
20 & 0.82 [0.77,0.86] & 0.63 [0.53,0.72] & 0.74 & 0.92 & 0.012 / 0.025 \\
22 & 0.63 [0.58,0.67] & 0.53 [0.43,0.63] & 0.58 & 0.92 & 0.009 / 0.025 \\
23 & 0.86 [0.81,0.90] & 0.79 [0.71,0.87] & 0.47 & 0.92 & 0.012 / 0.018 \\
25 & 0.78 [0.73,0.82] & 0.84 [0.77,0.91] & 0.50 & 0.91 & 0.012 / 0.013 \\
28 & 0.94 [0.89,0.98] & 0.98 [0.96,0.99] & 0.76 & 0.91 & 0.009 / 0.004 \\
\bottomrule

%% file: tab_layers_full.tex
8 & 0.22 & 0.39 & 0.43 \\
9 & 0.41 & 0.75 & 0.41 \\
10 & 0.52 & 0.55 & 0.27 \\
11 & 0.18 & 0.39 & 0.24 \\
12 & 0.77 & 0.52 & 0.17 \\
13 & 0.28 & 0.28 & 0.52 \\
14 & 0.38 & 0.61 & 0.74 \\
15 & 0.60 & 0.53 & 0.67 \\
16 & 0.44 & 0.41 & 0.44 \\
17 & 0.73 & 0.61 & 0.55 \\
18 & 0.90 & 0.60 & 0.52 \\
19 & 0.89 & 0.91 & 0.86 \\
20 & 0.82 & 0.63 & 0.74 \\
21 & 0.80 & 0.48 & 0.52 \\
22 & 0.63 & 0.53 & 0.58 \\
23 & 0.86 & 0.79 & 0.47 \\
24 & 0.72 & 0.75 & 0.53 \\
25 & 0.78 & 0.84 & 0.50 \\
26 & 0.81 & 0.80 & 0.65 \\
27 & 0.86 & 0.77 & 0.74 \\
28 & 0.94 & 0.98 & 0.76 \\
\bottomrule